\documentclass{article}
\usepackage{ijcai26}

\usepackage{xcolor}
\usepackage{times}
\usepackage{soul}
\usepackage{url}
\usepackage[hidelinks]{hyperref}
\usepackage[utf8]{inputenc}
\usepackage[small]{caption}
\usepackage{graphicx}
\usepackage{amsmath}
\usepackage{amsthm}
\usepackage{booktabs}
\usepackage{algorithm}
\usepackage{algorithmic}
\usepackage[switch]{lineno}
\usepackage{booktabs}
\usepackage{multirow}
\usepackage{arydshln}
\usepackage{float}
\usepackage{hyperref} 

\usepackage[T1]{fontenc}
\usepackage[utf8]{inputenc}
\usepackage{tikz}
\usetikzlibrary{arrows.meta, positioning, calc}

\usepackage{listings}
\definecolor{stageblue}{RGB}{41,98,194}
\definecolor{stagefill}{RGB}{224,234,252}
\definecolor{configamber}{RGB}{180,100,0}
\definecolor{configfill}{RGB}{255,243,218}
\definecolor{exgreen}{RGB}{15,120,60}
\definecolor{exfill}{RGB}{220,245,230}
\definecolor{arrowgray}{RGB}{70,70,70}
\definecolor{convpurple}{RGB}{100,40,160}
\definecolor{convfill}{RGB}{240,228,255}

\tikzset{
  stage/.style={
    draw=stageblue, fill=stagefill,
    rounded corners=3pt,
    minimum width=3.6cm, minimum height=0.85cm,
    align=center, font=\footnotesize,
    line width=0.8pt, inner sep=4pt
  },
  config/.style={
    draw=configamber, fill=configfill,
    rounded corners=2pt,
    minimum width=1.9cm, minimum height=0.52cm,
    align=center, font=\tiny\ttfamily,
    line width=0.6pt, inner sep=3pt
  },
  example/.style={
    draw=none, fill=none,
    text width=3cm, align=left,
    font=\scriptsize\itshape\color{exgreen}, inner sep=2pt
  },
  mainflow/.style={
    ->, >=Stealth,
    line width=0.8pt, color=arrowgray
  },
  configflow/.style={
    ->, >=Stealth,
    line width=0.5pt, color=configamber, dashed
  },
  exlink/.style={
    -, line width=0.0pt  
  },
  convscript/.style={
    draw=convpurple, fill=convfill,
    rounded corners=2pt,
    minimum width=1.9cm, minimum height=0.52cm,
    align=center, font=\small,
    line width=0.9pt, inner sep=6pt
  },
  convflow/.style={
    ->, >=Stealth,
    line width=0.5pt, color=convpurple
  },
}

\title{Policy-as-Logic for Robust Reasoning over Rules}

\author{
Rahul Nair$^1$
\and
Bastian Lipka$^2$\and
Elizabeth Daly$^1$
\affiliations
$^1$IBM Research\\
$^2$IBM\\
\emails
\{rahul.nair@ie.ibm.com,
lipka.bastian@ibm.com,
elizabeth.daly@ie.ibm.com\}
}

\begin{document}

\maketitle

\begin{abstract}
In many practical applications of generative AI systems, from tax rules to airline baggage allowance, responses to natural language queries must respect written policies or rules. We present a hybrid symbolic approach that expresses policies in formal logic and at inference time exploits the representation power of language models for fact extraction to ground predicates, and an answer set solver for reasoning such that responses are interpretable, auditable, and as we show, accurate and robust under input perturbations. Specifically, we show this separation of extraction and reasoning steps outperforms policy-as-prompt and policy-as-code methods in most cases with \textasciitilde10x reduction in token usage. The results point to the value of structured reasoning and symbolic solvers in conjunction with generative models to make robust decisions involving objective criteria.
\end{abstract}

\section{Introduction}
\label{sec:intro}
Many institutional processes are governed by written policies that specify how decisions should be made. Explicit written policies arise in a broad range of domains like safety, content moderation, business processes such as taxation, pricing, medical risk assessment, among others. Answering user queries reliably against these policies depend on human review which remains a practical reality for many critical domains. Automation of this decision making involves reasoning over the policy text and faithfully answering user queries \cite{palla2025policy}. When large language models (LLMs) are involved to facilitate this automation, decision making is no longer robust and subject to the stochastic nature of generative models. Subtle variations in input or policy text, for example tonal shifts or reordering of policy rules, can change the outcome ~\cite{ye2024justice,zhu2024promptbench}, making end-to-end LLM approaches unreliable.

Policy-based decision making involves two distinct capabilities: understanding the context of a user query and how it maps to a policy, and second reasoning over the policy to arrive at a consequence. While LLMs exploit their representation power for encoding context well, their reasoning capabilities involving discrete entities is unreliable \cite{zhou2025rulearena}. 

We further distinguish between rules arising from objective criteria or \emph{knowledge} (e.g. the weight of a bag when deciding a baggage fee) versus those that are subjective, or rely on \emph{belief} (e.g. a message intending to cause harm in a content moderation policy). In the latter, we are reliant on a model's view of the world to assess harm. Real world policies exist in a spectrum between objective and subjective criteria. 

Our Policy-as-logic (PaL) proposal is to express policy in formal logic. We use Answer Set Programs (ASPs) an extension of first order logic that allow for reasoning over defaults. Through the strict separation of fact extraction using LLMs and reasoning using classical solvers, we aim to improve robustness of pipelines involving policies. To study robustness, we systematically perturb inputs and observe decision consistency. Using two benchmark datasets we show significant performance gains at a fraction of the token cost when using formal logic for objective policies. For subjective policies however the benefit of reasoning is limited on account of extraction errors.


This work makes two contributions. First, we propose \emph{policy-as-logic} for automated decision making over rules and show its applicability in a broad range of practical domains. Second, we empirically demonstrate our methods over systematic input perturbations that structured reasoning methods such as ours are robust.

\section{Related Work}
\label{sec:background}

There is a large literature on the use of logic and large language models, see \cite{liu2025logical} for a recent survey. Neurosymbolic approaches like LINC \cite{olausson2023linc} look to derive first order logic from inputs and use theorem provers for reasoning. \cite{yang2023coupling} similarly use LLMs for semantic parsing and answer set programs for reasoning, although in their case the program is hand crafted. They use their LLM-ASP pipeline for several natural language tasks and appear to be the first to marry the parsing power of LLMs with declarative nature of answer set programs. \cite{pan-etal-2023-logic} additionally add self-correcting modules to improve error rates to a similar pipeline. They focus exclusively on logical reasoning tasks. In contrast, we study real-world policy reasoning. \cite{hoveyda2026orlog} apply probabilistic reasoning to address more specialized queries that may include exclusions, negations and other structural properties, which follows previous work on translating text to probabilistic programs \cite{wong2023word}.

In LLM-as-a-judge \cite{zheng2023judging} applications the criteria used to evaluate instances can also be considered a policy \cite{desmond2025evalassist}. The dominant paradigm to address these cases is \emph{policy-as-prompt}, where the policy language is used as context to LLM queries \cite{palla2025policy}. This is the used also in safeguards for LLMs, for example in GPT-OSS-Safeguards\footnote{\url{https://developers.openai.com/cookbook/articles/gpt-oss-safeguard-guide}} and Granite Guardian \cite{padhi2024graniteguardian}. 

Another approach is via \emph{policy-as-code} where policies and inputs are translated to executable programs \cite{dou2026deonticbench,yang2023coupling}. Generative computing frameworks allow for 
more flexible constructions mixing solvers, programs, and LLM prompts.

Several benchmarks like RuleArena \cite{zhou2025rulearena}, PolyGuard \cite{kumar2025polyguard}, DeonticBench \cite{dou2026deonticbench}, CLBench \cite{dou2026cl} exist for evaluating policy compliance. For this paper, we experiment with RuleArena for very discrete states and Polyguard for policies that involve more conceptual atoms.


\section{Methods}
\label{sec:methods}

\begin{figure}[ht]
    \centering
    \resizebox{\columnwidth}{!}{
    \begin{tikzpicture}[node distance=0.7cm]

  \node[stage] (stage-input) {%
    \textbf{Input Query}%
  };

  \node[stage, below=of stage-input] (stage-extract) {%
    \textbf{LLM Extraction}\\[1pt]
    {\tiny query $\times$ schema $\rightarrow$ JSON facts}%
  };

  \node[stage, below=of stage-extract] (stage-convert) {%
    \textbf{Grounding}\\[1pt]
    {\tiny JSON facts $\rightarrow$ grounded atoms}%
  };

  \node[stage, below=of stage-convert] (stage-solve) {%
    \textbf{Solver}\\[1pt]
    {\tiny atoms $+$ rules $\rightarrow$ answer set}%
  };

  \node[stage, below=of stage-solve] (stage-interpret) {%
    \textbf{Interpretation}\\[1pt]
    {\tiny answer set $\rightarrow$ decision}%
  };

  \draw[mainflow] (stage-input)   -- (stage-extract);
  \draw[mainflow] (stage-extract) -- (stage-convert);
  \draw[mainflow] (stage-convert) -- (stage-solve);
  \draw[mainflow] (stage-solve)   -- (stage-interpret);


  \node[convscript, left=0.3cm of stage-input] (conv-script) {%
    \textbf{Semantic Parsing}
  };

  \node[config, above=0.45cm of conv-script] (policy) {%
    \textbf{Policy text}
  };
  
  \node[config, left=0.5cm of stage-extract]   (cfg-schema)   {\texttt{schema.json}};
  \node[config, left=0.5cm of stage-convert]   (cfg-logic)    {\texttt{logic\_config}};
  \node[config, left=0.5cm of stage-solve]     (cfg-policy)   {\texttt{policy.lp}};
  \node[config, left=0.5cm of stage-interpret] (cfg-decision) {\texttt{decision\_config}};

  \draw[configflow] (policy) -- (conv-script);
  \draw[convflow] ([xshift=-1cm]conv-script.south) to[out=270, in=180]
        ([xshift=-0.3cm]cfg-schema.west) -- (cfg-schema.west);

  \draw[convflow] ([xshift=-1.2cm]conv-script.south) to[out=270, in=180]
        ([xshift=-0.3cm]cfg-policy.west) -- (cfg-policy.west);

  \draw[configflow] (cfg-schema.east)   -- (stage-extract.west);
  \draw[configflow] (cfg-logic.east)    -- (stage-convert.west);
  \draw[configflow] (cfg-policy.east)   -- (stage-solve.west);
  \draw[configflow] (cfg-decision.east) -- (stage-interpret.west);

  \node[example, right=0.1cm of stage-input] (ex-input) {%
    ``I am travelling in Economy class with one 20 kg bag. Is this allowed?''%
  };

  \node[example, right=0.1cm of stage-extract] (ex-extract) {%
    \upshape\ttfamily\{pass\_class: economy,\\
    bag\_weight: 20\}%
  };

  \node[example, right=0.1cm of stage-convert] (ex-convert) {%
    \upshape\ttfamily passenger\_class(economy).\\
    num\_checked\_bags(1).\\
    checked\_bag\_weight(1,200).\\
    \itshape\rmfamily{\tiny(20\,kg $\times$ 10)}%
  };

  \node[example, right=0.1cm of stage-solve] (ex-solve) {%
    \upshape\ttfamily decision(yes).\\
    total\_fee(0).%
  };

  \node[example, right=0.1cm of stage-interpret] (ex-interpret) {%
    \upshape\ttfamily decision(yes) \rmfamily$\rightarrow$
    \textbf{\textcolor{exgreen}{YES}}%
  };

\end{tikzpicture}}
    \caption{Pipeline overview}
    \label{fig:overview}
\end{figure}
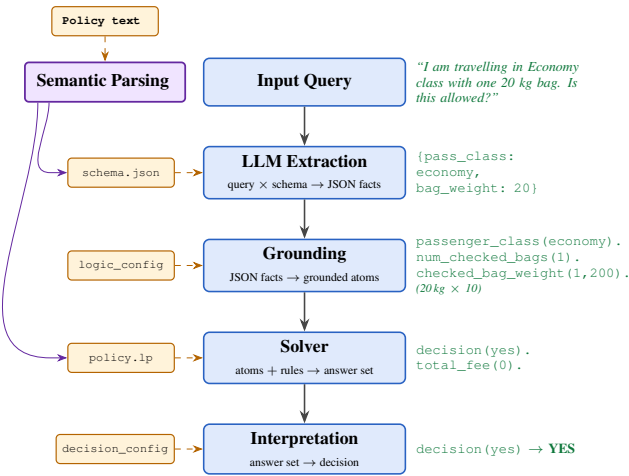

Given a policy $\mathcal{P}$ that describes some discrete outcomes of interest $y$, and a user query $x$, we are interested in a function $f$, which maps the user query to an outcome, i.e. $y=f(x, \mathcal{P})$. The input query $x$ is assumed to be in natural language as is the policy $\mathcal{P}$.

In the first semantic parsing step, we derive $P$, a logic program whose predicates encode facts, conditions, exceptions, and outcomes of the policy $\mathcal{P}$. This is performed once. At inference, for a query $x$, predicates involving variables are estimated using an LLM. The entire program $P$ is converted to a variable-free, propositional form through grounding. An answer set (or stable model) $M$ is a set of admissible decision outcomes of this grounded program. This is determined using a ASP solver. Figure \ref{fig:overview} shows the involved steps with an abridged example.

For $P$ we use a non-monotonic logic which has flexibility for practical applications as there may not be explicit rules to cover all possible outcomes. Additionally, they permit reasoning over defaults, such that when input queries are incomplete and result in partial grounding, the reasoning step yields the most likely outcomes. 

\paragraph{Semantic parsing} Using an LLM and a policy document, we first generate an answer set program which encode the policy. For our experiments we use Claude Opus 4.7 using a prompt shown in Appendix \ref{app:asp-prompt}. 
Note that this LLM-based translation does not guarantee complete policy coverage, and some of the domains in our experiments have known gaps, consistent with prior findings on LLM-to-ASP translation~\cite{ishay2023leveragingllmsforasp}. This step additionally generates a schema that is needed from the fact extractor and mappings that translate between text and logic and between logical consequence and decisions.

\paragraph{Extraction} Next, input queries are translated to structured output by prompting an LLM with the schema. The extraction step returns a list of facts in JSON. The LLM does not see the full policy text and only sees the schema.

\paragraph{Grounding} The extracted facts are translated to atoms using the mapping generated by the first step. These mappings handles translation between facts and atom names and handles types, e.g. from boolean facts to propositional atoms, and custom handlers for complex lists. The end result is a grounded ASP, i.e. a program in propositional logic without any variables.

\paragraph{Solver} Using Clingo \cite{gebser2014clingo,gebser2019multi} as the ASP solver we solve the grounded ASP to determine all stable models or answer sets. In some programs, optimization directives are needed for rule application ambiguity to ensure uniqueness of solutions. 

\paragraph{Interpretation} Finally, the output atoms are mapped back to domain decisions. Mapping tasks are deterministic and depend on the domain. For numeric values this can involve scaling (e.g. from cents to dollars), for categorical variables can involve standardization, or use structured objects as decisions.

Two points on computation are worth noting. Grounding can potentially result in an exponentially large search space. As we ground the program for each input query our variable expansion typically small, so the search space is in the order of number of predicates in $P$. Secondly, the resulting answer set may not be unique and a policy under a query may allow for multiple outcomes. In such cases, the recommended decision can be selected based on context such as picking the least cost decision or optimizing other criteria. In our experiments $|M|$ turns out to be small (see Table \ref{tab:answer-set-distribution}). Our method therefore does not introduce any latencies beyond the LLM call.

\paragraph{Evaluation} Our primary evaluation measure is accuracy (Acc) which denotes the fraction of cases where the outcome decisions match ground truth. To evaluate robustness, we make local perturbations of input queries and report accuracy across all perturbations (Rob). Following \cite{ye2024justice} we apply six language reformulating perturbations to every query: verbosity, paraphrase, distraction, misleading context, cheerful sentiment, and frustrated sentiment. The perturbations are generated by an LLM. A secondary validation step using an llm-as-a-judge to evaluate if the semantic meaning of the perturbation is preserved. 


\paragraph{Baselines} The competing paradigms of \emph{policy-as-prompt} \cite{palla2025policy} and \emph{policy-as-code} \cite{dou2026deonticbench} serve as baselines to our proposals.

\section{Experiments}

\subsection{Setup}
We validate our approach across four domains from two benchmarks. The Airline, Tax, and NBA domains originate from RuleArena~\cite{zhou2025rulearena}; the HR domain originates from the PolyGuard training set~\cite{kumar2025polyguard}. Each domain involves the following tasks:

\begin{enumerate}
    \item \emph{Airline baggage fees}: Requires calculating the total cost for one or multiple passengers, consisting of the flight ticket and the checked baggage fees. 
    \item \emph{Income tax}: Requires calculating the income tax for a given person or family based on their financial situation. 
    \item \emph{NBA Transactions}: Requires detecting illegal salary violations in player transactions and identifying the rule-violating team and operation.
    \item \emph{HR content moderation}: Requires classifying workplace situations as safe or unsafe according to an HR policy.
\end{enumerate}

The first three domains involve predominantly objective, knowledge-based criteria. The HR domain involves subjective, belief-based criteria and serves as a boundary case. For the \emph{policy-as-prompt} baseline, we test 0-shot and 1-shot using prompts from the benchmark \cite{zhou2025rulearena} with no and one in context exemplars. We compare with various open weight LLMs namely: GPT-OSS-120B, Qwen-2.5 72B, Llama 3.3 70B and Granite-4.1 8B. For \emph{policy-as-code} we compare with numbers reported in \cite{dou2026deonticbench} as the code was released during this writing. They report metrics just for the airline case (see accuracy values in Table 2 which range from $0.0$ for Kimi K2 to $0.40$ for GPT-5.1). Additionally \cite{zhou2025rulearena} report partial numbers for code generation on the airline domain (see accuracy values in Table 10 which range from $0.18$ for LLama-70B, level 2 to $0.44$ for GPT-4o, Level 1). 

In our experiments we treat all decisions to be categorical. So accuracy implies an exact match to the ground truth. This is a higher threshold to meet for continuous valued decisions like bag fees. All robustness measures are based on six perturbations to each input query. 

\subsection{Results}
Table \ref{tab:headline-results} shows the main results with results by difficulty level in the appendix in Table \ref{tab:main-results}.

\begin{table}[!h]
\centering
\caption{Accuracy $\uparrow$ and Robustness $\uparrow$ across four domains and four LLMs, aggregated
over all queries per domain (Airline: $n{=}300$, Tax: $n{=}300$,
NBA: $n{=}216$, HR: $n{=}300$). HR has no 1-shot setting.}
\label{tab:headline-results}
\scriptsize
\setlength{\tabcolsep}{4pt}
\begin{tabular}{@{}ll cc cc cc cc@{}}
\toprule
\multirow{2}{*}{\textbf{Models}} & \multirow{2}{*}{\textbf{Method}} & \multicolumn{2}{c}{\textbf{Airline}} & \multicolumn{2}{c}{\textbf{Tax}} & \multicolumn{2}{c}{\textbf{NBA}} & \multicolumn{2}{c}{\textbf{HR}} \\
\cmidrule(lr){3-4} \cmidrule(lr){5-6} \cmidrule(lr){7-8} \cmidrule(lr){9-10}
 &  & Acc & Rob & Acc & Rob & Acc & Rob & Acc & Rob \\
\midrule
GPT-OSS  & 0-shot         & 0.38 & 0.27 & 0.07 & 0.00 & 0.25 & 0.22 & 0.95 & \textbf{0.94} \\
120B  & 1-shot         & 0.37 & 0.17 & 0.10 & 0.00 & 0.12 & 0.13 & --- & --- \\
  & PaL (ours) & \textbf{1.00} & \textbf{0.98} & \textbf{0.31} & \textbf{0.29} & \textbf{0.48} & \textbf{0.46} & \textbf{0.96} & 0.93 \\
\hdashline
Qwen-2.5  & 0-shot         & 0.01 & 0.01 & 0.02 & 0.00 & 0.39 & 0.39 & \textbf{0.96} & \textbf{0.94} \\
 72B & 1-shot         & 0.07 & 0.04 & 0.07 & 0.00 & 0.33 & 0.36 & --- & --- \\
  & PaL (ours)& \textbf{0.94} & \textbf{0.93} & \textbf{0.31} & \textbf{0.29} & \textbf{0.50} & \textbf{0.48} & 0.94 & 0.93 \\
\hdashline
Llama-3.3  & 0-shot         & 0.01 & 0.01 & 0.00 & 0.00 & 0.27 & 0.26 & 0.96 & 0.93 \\
70B  & 1-shot         & 0.07 & 0.04 & 0.03 & 0.00 & 0.32 & 0.32 & --- & --- \\
  & PaL (ours)& \textbf{0.97} & \textbf{0.94} & \textbf{0.31} & \textbf{0.24} & \textbf{0.49} & \textbf{0.47} & \textbf{0.97} & \textbf{0.94} \\
\hdashline
 Granite-4.1 & 0-shot         & 0.01 & 0.00 & 0.00 & 0.00 & 0.32 & 0.34 & \textbf{0.96} & \textbf{0.95} \\
 8B & 1-shot         & 0.01 & 0.00 & 0.00 & 0.00 & 0.29 & 0.31 & --- & --- \\
  & PaL (ours)& \textbf{0.61} & \textbf{0.62} & \textbf{0.31} & \textbf{0.26} & \textbf{0.36} & \textbf{0.37} & 0.93 & 0.90 \\
\bottomrule
\end{tabular}
\end{table}

\paragraph{Performance} On Airline, our method achieves accuracy numbers between 0.94 and 1.00 across all models, while baselines don't exceed 0.38 for policy-as-prompt and 0.40 for policy-as-code. The smallest model, Granite-4.1 8B, reaches 0.61 with the pipeline compared to 0.01 without. On Tax, all baselines remain below 0.10 while PaL scores 0.31 across all models. On the NBA domain, the pipeline outperforms baselines, though the gap is smaller suggesting multi-operation scenarios place a higher burden on the extraction quality. 

\paragraph{Robustness} Robustness follows a similar pattern but widens the gap. On Tax, every baseline across all four models drops to 0.00 robustness, meaning that not a single perturbed query is answered correctly. PaL maintains robustness close to the accuracy across all four domains. This is a direct consequence of the architecture: since the grounding, solving, and interpretation step is deterministic. The only source of robustness loss is extraction quality  under perturbations.

\paragraph{Conceptual atoms} On HR, the pattern reverses. The pipeline offers no systematic advantage over the baseline. PaL and the 0-shot baseline score between 0.93 and 0.96. On Qwen and Granite the baseline outperforms the pipeline. The HR rules are  simple (e.g., ``Do not work under the influence of drugs or alcohol.''), so the solver contributes no logical reasoning beyond what the LLM does in a single forward pass. The compliance classification in this domain requires conceptual decision making rather than logical reasoning. This result supports our claim that the robustness gains on Airline, Tax, and NBA originate from the architectural separation of natural language understanding and logical reasoning.

\paragraph{Token efficiency} We additionally compare the number of tokens for each method per query. As the entire policy forms a part of the context for direct queries and policy-as-logic needs only the schema, our method needs fewer tokens by an order of magnitude in most domains as shown in Table \ref{tab:token-efficiency}.

\begin{table}[t]
\centering
\small
\caption{Per-query token usage for the pipeline vs the 
LLM-only baselines. Counts use the \texttt{o200k\_base} tokenizer applied
uniformly to both methods. \textbf{In} = prompt tokens; \textbf{Out} = response tokens.}
\label{tab:token-efficiency}
\begin{tabular}{llrrr}
\toprule
Domain & Method &  In (mean) & Out (mean) & Total \\
\midrule
Airline & 0-shot &  11{,}620 &  63 & 11{,}684 \\
 (n=300) & 1-shot &  12{,}836 &  67 & 12{,}903 \\
 & PaL (ours)       &     895 & 280 &  1{,}175 \\
\midrule
 Tax & 0-shot &   9{,}788 &  28 &  9{,}816 \\
 (n=300) & 1-shot &  17{,}047 &  25 & 17{,}072 \\
 & PaL (ours)       &   3{,}537 & 606 &  4{,}143 \\
\midrule
 NBA & 0-shot &  21{,}793 &  53 & 21{,}846 \\
 (n=216) & 1-shot &  23{,}663 &  35 & 23{,}698 \\
 & PaL (ours)       &  2{,}262 & 617 &  2{,}879 \\
\midrule
 HR & 0-shot &     468 &  34 &    502 \\
 (n=300) & PaL (ours)       &     904 &  71 &    975 \\
\bottomrule
\end{tabular}
\end{table}

\paragraph{Size of answer sets} In our setup the solver returns a set of answers that are compatible with the grounded predicates. Since a written policy is intended to guide decision making, it should map a well specified query to a specific outcome. However, in some cases there is ambiguity in how rules are meant to be applied. For instance in the airline domain, which bag is considered complimentary can impact total fees if a passenger has multiple  with different weights. For this case, we add an optimization directive to the ASP to pick the assignment with the lowest cost. Table \ref{tab:answer-set-distribution} shows the cardinality of the answer sets showing that the feasible answers are no larger than $3$. Similar ambiguity does not exist in the other tested domains.

\begin{table}[ht]
\centering
\small
\caption{Size of answer sets returned by domain.}
\label{tab:answer-set-distribution}
\begin{tabular}{lrrrrrrrr}
\toprule
Domain            & $N$ & $|M|=1$ & 2 & 3 & Min & Max & Mean \\
\midrule
Airline & 300 & 206 & 64 & 30 & 1 & 3 & 1.41 \\
Tax     & 300 & 300 & 0  & 0  & 1 & 1 & 1.00 \\
NBA     & 216 & 216 & 0  & 0  & 1 & 1 & 1.00 \\
HR      & 300 & 300 & 0  & 0  & 1 & 1 & 1.00 \\
\bottomrule
\end{tabular}
\end{table}

\section{Discussion}

We have presented policy-as-logic formulation that had broad applicability in practice for domains where there are written rules. Preliminary evidence from a few representative domains suggests significant performance gains where  automated decision pipelines leverage structured reasoning. In our experiments, gains are predominantly in domains where rules are based on knowledge rather than beliefs. When the policy relies on belief, the extraction step must make the judgment call, and the solver adds less value. In other words, policies expressed in objective measures are better suited to be encoded in formal logic compared to those relying on subjective criteria. 

We envision such methods can be part of broader agentic systems or specialized tool calls, where input queries that have been adjudged to be in scope for a specific policy can be routed to structured reasoning solvers similar to those proposed herein. If the context provided by users is partial and missing key attributes, interactive human-in-the-loop can be used to capture missing information. 

Several performance improvements are viable as next steps. For more nuanced criteria, instead of schema based fact extraction that attempts to perform it in one shot, we can disaggregate calls and leverage llm-as-a-judge literature to address problems of positional bias \cite{shi2025judging} and criteria phrasing \cite{desmond2025evalassist} to make fact extraction more grounded. More robust error analysis that goes from solver decisions via propositional atoms back to the original text would aid in diagnosis of error cases. Our results add to the evidence reported in other works\cite{dou2026deonticbench} that numerical errors are rife when applying complex business policies to queries.

\bibliographystyle{named}
\bibliography{ijcai26}

@inproceedings{palla2025policy,
  title={Policy-as-prompt: Rethinking content moderation in the age of large language models},
  author={Palla, Konstantina and Garc{\'\i}a, Jos{\'e} Luis Redondo and Hauff, Claudia and Fabbri, Francesco and Damianou, Andreas and Lindstr{\"o}m, Henrik and Taber, Dan and Lalmas, Mounia},
  booktitle={Proceedings of the 2025 ACM Conference on Fairness, Accountability, and Transparency},
  pages={840--854},
  year={2025}
}

@article{liu2025logical,
  title={Logical reasoning in large language models: A survey},
  author={Liu, Hanmeng and Fu, Zhizhang and Ding, Mengru and Ning, Ruoxi and Zhang, Chaoli and Liu, Xiaozhang and Zhang, Yue},
  journal={arXiv preprint arXiv:2502.09100},
  year={2025}
}

@article{dou2026cl,
  title={CL-bench: A Benchmark for Context Learning},
  author={Dou, Shihan and Zhang, Ming and Yin, Zhangyue and Huang, Chenhao and Shen, Yujiong and Wang, Junzhe and Chen, Jiayi and Ni, Yuchen and Ye, Junjie and Zhang, Cheng and others},
  journal={arXiv preprint arXiv:2602.03587},
  year={2026}
}

@misc{padhi2024graniteguardian,
      title={Granite Guardian}, 
      author={Inkit Padhi and Manish Nagireddy and Giandomenico Cornacchia and Subhajit Chaudhury and Tejaswini Pedapati and Pierre Dognin and Keerthiram Murugesan and Erik Miehling and Martín Santillán Cooper and Kieran Fraser and Giulio Zizzo and Muhammad Zaid Hameed and Mark Purcell and Michael Desmond and Qian Pan and Zahra Ashktorab and Inge Vejsbjerg and Elizabeth M. Daly and Michael Hind and Werner Geyer and Ambrish Rawat and Kush R. Varshney and Prasanna Sattigeri},
      year={2024},
      eprint={2412.07724},
      archivePrefix={arXiv},
      primaryClass={cs.CL},
      url={https://arxiv.org/abs/2412.07724}, 
}

@inproceedings{desmond2025evalassist,
  title={Evalassist: Llm-as-a-judge simplified},
  author={Desmond, Michael and Ashktorab, Zahra and Geyer, Werner and Daly, Elizabeth M and Cooper, Martin Santillan and Pan, Qian and Nair, Rahul and Wagner, Nico and Pedapati, Tejaswini},
  booktitle={Proceedings of the AAAI Conference on Artificial Intelligence},
  volume={39},
  number={28},
  pages={29637--29639},
  year={2025}
}

@article{wong2023word,
  title={From word models to world models: Translating from natural language to the probabilistic language of thought},
  author={Wong, Lionel and Grand, Gabriel and Lew, Alexander K and Goodman, Noah D and Mansinghka, Vikash K and Andreas, Jacob and Tenenbaum, Joshua B},
  journal={arXiv preprint arXiv:2306.12672},
  year={2023}
}

@inproceedings{hoveyda2026orlog,
  title={OrLog: Resolving Complex Queries with LLMs and Probabilistic Reasoning},
  author={Hoveyda, Mohanna and Piepenbrock, Jelle and de Vries, Arjen P and de Rijke, Maarten and Hasibi, Faegheh},
  booktitle={European Conference on Information Retrieval},
  pages={98--114},
  year={2026},
  organization={Springer}
}

@inproceedings{zhou2025rulearena,
  title={Rulearena: A benchmark for rule-guided reasoning with llms in real-world scenarios},
  author={Zhou, Ruiwen and Hua, Wenyue and Pan, Liangming and Cheng, Sitao and Wu, Xiaobao and Yu, En and Wang, William Yang},
  booktitle={Proceedings of the 63rd Annual Meeting of the Association for Computational Linguistics (Volume 1: Long Papers)},
  pages={550--572},
  year={2025}
}

@inproceedings{shi2025judging,
  title={Judging the judges: A systematic study of position bias in llm-as-a-judge},
  author={Shi, Lin and Ma, Chiyu and Liang, Wenhua and Diao, Xingjian and Ma, Weicheng and Vosoughi, Soroush},
  booktitle={Proceedings of the 14th International Joint Conference on Natural Language Processing and the 4th Conference of the Asia-Pacific Chapter of the Association for Computational Linguistics},
  pages={292--314},
  year={2025}
}

@article{dou2026deonticbench,
  title={DeonticBench: A Benchmark for Reasoning over Rules},
  author={Dou, Guangyao and Brena, Luis and Deo, Akhil and Jurayj, William and Zhang, Jingyu and Holzenberger, Nils and Van Durme, Benjamin},
  journal={arXiv preprint arXiv:2604.04443},
  year={2026}
}

@article{kumar2025polyguard,
  title={Polyguard: A multilingual safety moderation tool for 17 languages},
  author={Kumar, Priyanshu and Jain, Devansh and Yerukola, Akhila and Jiang, Liwei and Beniwal, Himanshu and Hartvigsen, Thomas and Sap, Maarten},
  journal={arXiv preprint arXiv:2504.04377},
  year={2025}
}

@article{zheng2023judging,
  title={Judging llm-as-a-judge with mt-bench and chatbot arena},
  author={Zheng, Lianmin and Chiang, Wei-Lin and Sheng, Ying and Zhuang, Siyuan and Wu, Zhanghao and Zhuang, Yonghao and Lin, Zi and Li, Zhuohan and Li, Dacheng and Xing, Eric and others},
  journal={Advances in neural information processing systems},
  volume={36},
  pages={46595--46623},
  year={2023}
}

@inproceedings{yang2023coupling,
  title={Coupling large language models with logic programming for robust and general reasoning from text},
  author={Yang, Zhun and Ishay, Adam and Lee, Joohyung},
  booktitle={Findings of the association for computational linguistics: ACL 2023},
  pages={5186--5219},
  year={2023}
}

@inproceedings{ishay2023leveragingllmsforasp,
    title     = {{Leveraging Large Language Models to Generate Answer Set Programs}},
    author    = {Ishay, Adam and Yang, Zhun and Lee, Joohyung},
    booktitle = {{Proceedings of the 20th International Conference on Principles of Knowledge Representation and Reasoning}},
    pages     = {374--383},
    year      = {2023},
    month     = {8},
    doi       = {10.24963/kr.2023/37},
    url       = {https://doi.org/10.24963/kr.2023/37},
}

@inproceedings{pan-etal-2023-logic,
    title = "Logic-{LM}: Empowering Large Language Models with Symbolic Solvers for Faithful Logical Reasoning",
    author = "Pan, Liangming  and
      Albalak, Alon  and
      Wang, Xinyi  and
      Wang, William",
    booktitle = "Findings of the Association for Computational Linguistics: EMNLP 2023",
    month = dec,
    year = "2023",
    publisher = "Association for Computational Linguistics",
    url = "https://aclanthology.org/2023.findings-emnlp.248/",
    doi = "10.18653/v1/2023.findings-emnlp.248",
    pages = "3806--3824",
}

@inproceedings{olausson2023linc,
  title={LINC: A neurosymbolic approach for logical reasoning by combining language models with first-order logic provers},
  author={Olausson, Theo and Gu, Alex and Lipkin, Ben and Zhang, Cedegao and Solar-Lezama, Armando and Tenenbaum, Joshua and Levy, Roger},
  booktitle={Proceedings of the 2023 Conference on Empirical Methods in Natural Language Processing},
  pages={5153--5176},
  year={2023}
}

@article{gebser2019multi,
  title={Multi-shot ASP solving with clingo},
  author={Gebser, Martin and Kaminski, Roland and Kaufmann, Benjamin and Schaub, Torsten},
  journal={Theory and Practice of Logic Programming},
  volume={19},
  number={1},
  pages={27--82},
  year={2019},
  publisher={Cambridge University Press}
}

@article{gebser2014clingo,
  title={Clingo= ASP+ control: Preliminary report},
  author={Gebser, Martin and Kaminski, Roland and Kaufmann, Benjamin and Schaub, Torsten},
  journal={arXiv preprint arXiv:1405.3694},
  year={2014}
}

@article{ye2024justice,
  title={Justice or Prejudice? Quantifying Biases in LLM-as-a-Judge},
  author={Ye, Jiayi and Wang, Yanbo and Huang, Yue and Chen, Dongping and Zhang, Qihui and Moniz, Nuno and Gao, Tian and Geyer, Werner and Huang, Chao and Chen, Pin-Yu and Chawla, Nitesh V. and Zhang, Xiangliang},
  journal={arXiv preprint arXiv:2410.02736},
  year={2024}
}

@article{zhu2024promptbench,
  title={Promptbench: A unified library for evaluation of large language models},
  author={Zhu, Kaijie and Zhao, Qinlin and Chen, Hao and Wang, Jindong and Xie, Xing},
  journal={Journal of Machine Learning Research},
  volume={25},
  number={254},
  pages={1--22},
  year={2024}
}

\section{Appendices}

\subsection{Policy-to-ASP Generation Prompt}
\label{app:asp-prompt}

The following prompt template is used to generate the ASP rule generation from a textual policy document. The template takes three inputs: \texttt{\{policy\_text\}}, the full policy document in natural language; \texttt{\{output\_description\}}, a short description of the expected decision type (e.g., ``a numeric total baggage cost in USD'' or ``a categorical decision: SAFE or UNSAFE''); and \texttt{\{domain\_notes\}}, optional domain-specific instructions (e.g., ``dollar amounts may exceed 32-bit integers, use cent-scaling'').

\begin{small}
\begin{lstlisting}
You are an expert in Answer Set Programming (ASP)
using the Clingo solver (v5.8.0). Your task is to
translate a natural-language policy document into a
complete, executable ASP program (policy.lp).

== POLICY DOCUMENT ==
{policy_text}

== DESIRED OUTPUT ==
{output_description}

== ADDITIONAL NOTES ==
{domain_notes}

== REQUIREMENTS ==

1. Encode every rule, exception, and edge case in
   the policy document. Use default negation (not)
   where the policy implies a default outcome when
   no explicit rule applies.

2. Do not hardcode any facts that come from user
   queries. The program must be generic. Input facts
   will be provided as ground atoms at inference
   time.

3. Use descriptive snake_case predicate names. Add
   comment blocks explaining which part of the
   policy each section encodes.

4. Use only standard Clingo constructs: choice
   rules, aggregates, weak constraints. Clingo
   integers are 32-bit. If large numbers are
   involved, scale them (e.g., cents instead of
   dollars) and document the scale factor.

5. End with #show directives for all decision-
   relevant output atoms.

6. Return only the ASP code. No markdown fences,
   no explanations before or after.
\end{lstlisting}
\end{small}

\subsection{Extraction failures}
\label{app:extraction_errors}

Here we show some representative extraction failures using a small Granite-4.1 8B for each of the domains and the resulting change in decision. 

\subsubsection*{A.1 Airline: Default-class collapse}
\label{app:err-airline}

Query RA\_L0\_006 describes a passenger flying from Wuhan to Portland.
The schema's \texttt{route} field should reflect the international
origin region.

\smallskip\noindent\textbf{Extracted:}
\begin{lstlisting}
{ "route": "U.S.", ... }
\end{lstlisting}

\noindent\textbf{Correct:}
\begin{lstlisting}
{ "route": "China", ... }
\end{lstlisting}

Granite defaults to \texttt{"U.S."} regardless of origin city. The
misclassified route causes the solver to apply domestic instead of
international fees (\$2,249 vs.\ correct \$2,314). The same error
recurs on queries involving Toronto, Tokyo, and Nagoya.

\subsubsection*{A.2 Tax: Digit-level transcription error}
\label{app:err-tax}

Query RA\_TAX\_L2\_017 states gross receipts of \$29,314.

\smallskip\noindent\textbf{Extracted:}
\begin{lstlisting}
{ "gross_receipts": 29114, ... }
\end{lstlisting}

\noindent\textbf{Correct:}
\begin{lstlisting}
{ "gross_receipts": 29314, ... }
\end{lstlisting}

A single-digit substitution (3$\to$1). The \$200 input error propagates
through Schedule~C and self-employment tax to a \$1,956 difference in
the final decision.

\subsubsection*{A.3 NBA: Cross-field value confusion}
\label{app:err-nba}

Query RA\_NBA\_L1\_056 describes Player~A with a prior contract at
\$16M/year (2022) and a new signing operation at \$30M/year (2024--25),
both with Team~E.

\smallskip\noindent\textbf{Extracted:}
\begin{lstlisting}
{ "last_contract_annual_salary": 30000000,
  "last_contract_start_year": 2022, ... }
\end{lstlisting}

\noindent\textbf{Correct:}
\begin{lstlisting}
{ "last_contract_annual_salary": 16000000,
  "last_contract_start_year": 2022, ... }
\end{lstlisting}

Granite merges the salary from the new signing into the prior contract.
The inflated salary changes the maximum-salary classification, causing
the pipeline to miss a cap violation.

\subsubsection*{A.4 HR: Keyword trigger ignoring stance}
\label{app:err-hr}

Query SAFE\_0093 is an internal HR reminder: \emph{``Using substances
like vape pens before meetings violates our workplace policy.''}

\smallskip\noindent\textbf{Extracted:}
\begin{lstlisting}
{ "contains_substance_abuse_on_duty": true, ... }
\end{lstlisting}

\noindent\textbf{Correct:}
\begin{lstlisting}
{ "contains_substance_abuse_on_duty": false, ... }
\end{lstlisting}

The message prohibits substance use rather than admitting to it. Granite
triggers on surface keywords without modeling the author's stance,
flipping the decision from SAFE to UNSAFE.

\begin{table*}[t]
\centering
\caption{Accuracy and Robustness across four domains, four LLMs, and three difficulty levels. Robustness is the fraction of perturbed queries answered correctly, averaged across six perturbation types. Best value per model is in \textbf{bold}. PolyGuard HR has no difficulty levels.}
\label{tab:main-results}
\footnotesize
\setlength{\tabcolsep}{3.4pt}
\begin{tabular}{@{}ll rr rr rr rr@{}}
\toprule
\multirow{2}{*}{\textbf{Models}} & \multirow{2}{*}{\textbf{Settings}} & \multicolumn{2}{c}{\textbf{Level 0}} & \multicolumn{2}{c}{\textbf{Level 1}} & \multicolumn{2}{c}{\textbf{Level 2}} & \multicolumn{2}{c}{\textbf{All}} \\
\cmidrule(lr){3-4} \cmidrule(lr){5-6} \cmidrule(lr){7-8} \cmidrule(lr){9-10}
 &  & Acc & Rob & Acc & Rob & Acc & Rob & Acc & Rob \\
\midrule
\multicolumn{10}{c}{\textbf{Airline}} \\
\midrule
\multirow{3}{*}{GPT-OSS 120B}
  & 0-shot & 0.50 & 0.32 & 0.38 & 0.26 & 0.27 & 0.23 & 0.38 & 0.27 \\

  & 1-shot & 0.49 & 0.14 & 0.40 & 0.20 & 0.22 & 0.18 & 0.37 & 0.17 \\

  & \textbf{PaL} & \textbf{1.00} & \textbf{0.98} & \textbf{1.00} & \textbf{0.99} & \textbf{1.00} & \textbf{0.98} & \textbf{1.00} & \textbf{0.98} \\
\hdashline
\multirow{3}{*}{Qwen-2.5 72B}
  & 0-shot & 0.02 & 0.01 & 0.00 & 0.01 & 0.02 & 0.00 & 0.01 & 0.01 \\

  & 1-shot & 0.15 & 0.05 & 0.05 & 0.05 & 0.02 & 0.01 & 0.07 & 0.04 \\

  & \textbf{PaL} & \textbf{0.91} & \textbf{0.93} & \textbf{0.97} & \textbf{0.94} & \textbf{0.95} & \textbf{0.93} & \textbf{0.94} & \textbf{0.93} \\
\hdashline
\multirow{3}{*}{Llama-3.3 70B}
  & 0-shot & 0.04 & 0.01 & 0.00 & 0.00 & 0.00 & 0.01 & 0.01 & 0.01 \\

  & 1-shot & 0.17 & 0.10 & 0.03 & 0.02 & 0.01 & 0.00 & 0.07 & 0.04 \\

  & \textbf{PaL} & \textbf{0.97} & \textbf{0.94} & \textbf{0.96} & \textbf{0.96} & \textbf{0.98} & \textbf{0.94} & \textbf{0.97} & \textbf{0.94} \\
\hdashline
\multirow{3}{*}{Granite-4.1 8B}
  & 0-shot & 0.00 & 0.01 & 0.01 & 0.00 & 0.01 & 0.00 & 0.01 & 0.00 \\

  & 1-shot & 0.00 & 0.00 & 0.02 & 0.00 & 0.00 & 0.01 & 0.01 & 0.00 \\

  & \textbf{PaL} & \textbf{0.59} & \textbf{0.68} & \textbf{0.64} & \textbf{0.60} & \textbf{0.59} & \textbf{0.58} & \textbf{0.61} & \textbf{0.62} \\
\midrule
\multicolumn{10}{c}{\textbf{Tax}} \\
\midrule
\multirow{3}{*}{GPT-OSS 120B}
  & 0-shot & 0.21 & 0.00 & 0.01 & 0.00 & 0.00 & 0.00 & 0.07 & 0.00 \\

  & 1-shot & 0.29 & 0.00 & 0.00 & 0.00 & 0.00 & 0.00 & 0.10 & 0.00 \\

  & \textbf{PaL} & \textbf{0.58} & \textbf{0.56} & \textbf{0.28} & \textbf{0.24} & \textbf{0.07} & \textbf{0.05} & \textbf{0.31} & \textbf{0.29} \\
\hdashline
\multirow{3}{*}{Qwen-2.5 72B}
  & 0-shot & 0.03 & 0.00 & 0.02 & 0.00 & 0.00 & 0.00 & 0.02 & 0.00 \\

  & 1-shot & 0.21 & 0.00 & 0.01 & 0.00 & 0.00 & 0.00 & 0.07 & 0.00 \\

  & \textbf{PaL} & \textbf{0.58} & \textbf{0.56} & \textbf{0.28} & \textbf{0.24} & \textbf{0.07} & \textbf{0.05} & \textbf{0.31} & \textbf{0.29} \\
\hdashline
\multirow{3}{*}{Llama-3.3 70B}
  & 0-shot & 0.00 & 0.00 & 0.00 & 0.00 & 0.00 & 0.00 & 0.00 & 0.00 \\

  & 1-shot & 0.08 & 0.00 & 0.00 & 0.00 & 0.00 & 0.00 & 0.03 & 0.00 \\

  & \textbf{PaL} & \textbf{0.58} & \textbf{0.50} & \textbf{0.28} & \textbf{0.17} & \textbf{0.07} & \textbf{0.04} & \textbf{0.31} & \textbf{0.24} \\
\hdashline
\multirow{3}{*}{Granite-4.1 8B}
  & 0-shot & 0.00 & 0.00 & 0.00 & 0.00 & 0.00 & 0.00 & 0.00 & 0.00 \\

  & 1-shot & 0.00 & 0.00 & 0.00 & 0.00 & 0.00 & 0.00 & 0.00 & 0.00 \\

  & \textbf{PaL} & \textbf{0.58} & \textbf{0.49} & \textbf{0.28} & \textbf{0.22} & \textbf{0.07} & \textbf{0.05} & \textbf{0.31} & \textbf{0.26} \\
\midrule
\multicolumn{10}{c}{\textbf{NBA Transaction}} \\
\midrule
\multirow{3}{*}{GPT-OSS 120B}
  & 0-shot & 0.32 & 0.30 & 0.28 & 0.21 & 0.07 & 0.11 & 0.25 & 0.22 \\

  & 1-shot & 0.23 & 0.21 & 0.09 & 0.09 & 0.00 & 0.05 & 0.12 & 0.13 \\

  & \textbf{PaL} & \textbf{0.63} & \textbf{0.61} & \textbf{0.37} & \textbf{0.37} & \textbf{0.41} & \textbf{0.35} & \textbf{0.48} & \textbf{0.46} \\
\hdashline
\multirow{3}{*}{Qwen-2.5 72B}
  & 0-shot & 0.52 & 0.45 & 0.40 & 0.37 & 0.15 & 0.31 & 0.39 & 0.39 \\

  & 1-shot & 0.38 & 0.49 & 0.30 & 0.33 & 0.28 & 0.22 & 0.33 & 0.36 \\

  & \textbf{PaL} & \textbf{0.64} & \textbf{0.64} & \textbf{0.42} & \textbf{0.39} & \textbf{0.41} & \textbf{0.36} & \textbf{0.50} & \textbf{0.48} \\
\hdashline
\multirow{3}{*}{Llama-3.3 70B}
  & 0-shot & 0.38 & 0.34 & 0.23 & 0.24 & 0.15 & 0.15 & 0.27 & 0.26 \\

  & 1-shot & 0.49 & 0.48 & 0.20 & 0.24 & 0.24 & 0.18 & 0.32 & 0.32 \\

  & \textbf{PaL} & \textbf{0.64} & \textbf{0.63} & \textbf{0.38} & \textbf{0.37} & \textbf{0.41} & \textbf{0.36} & \textbf{0.49} & \textbf{0.47} \\
\hdashline
\multirow{3}{*}{Granite-4.1 8B}
  & 0-shot & 0.42 & 0.43 & 0.28 & 0.29 & 0.24 & \textbf{0.25} & 0.32 & 0.34 \\

  & 1-shot & \textbf{0.43} & 0.45 & 0.24 & 0.26 & 0.13 & 0.19 & 0.29 & 0.31 \\

  & \textbf{PaL} & \textbf{0.43} & \textbf{0.51} & \textbf{0.34} & \textbf{0.30} & \textbf{0.26} & 0.24 & \textbf{0.36} & \textbf{0.37} \\
\midrule
\multicolumn{10}{c}{\textbf{PolyGuard HR}} \\
\midrule
\multirow{2}{*}{GPT-OSS 120B}
  & 0-shot & --- & --- & --- & --- & --- & --- & 0.95 & \textbf{0.94} \\

  & \textbf{PaL} & --- & --- & --- & --- & --- & --- & \textbf{0.96} & 0.93 \\
\hdashline
\multirow{2}{*}{Qwen-2.5 72B}
  & 0-shot & --- & --- & --- & --- & --- & --- & \textbf{0.96} & \textbf{0.94} \\

  & \textbf{PaL} & --- & --- & --- & --- & --- & --- & 0.94 & 0.93 \\
\hdashline
\multirow{2}{*}{Llama-3.3 70B}
  & 0-shot & --- & --- & --- & --- & --- & --- & 0.96 & 0.93 \\

  & \textbf{PaL} & --- & --- & --- & --- & --- & --- & \textbf{0.97} & \textbf{0.94} \\
\hdashline
\multirow{2}{*}{Granite-4.1 8B}
  & 0-shot & --- & --- & --- & --- & --- & --- & \textbf{0.96} & \textbf{0.95} \\

  & \textbf{PaL} & --- & --- & --- & --- & --- & --- & 0.93 & 0.90 \\
\bottomrule
\end{tabular}
\end{table*}

\end{document}